\documentclass{ecai}
\usepackage{times}
\usepackage{graphicx}
\usepackage{latexsym}
\usepackage{tabularx}

\ecaisubmission 
\newcommand{\citet}[1]{\citeauthor{#1} \shortcite{#1}}

\usepackage{amsmath,amssymb}
\usepackage{xargs}
\usepackage[colorinlistoftodos,prependcaption]{todonotes}
\usepackage{subfigure}

\usepackage{hyperref}

\usepackage{wasysym} % \smiley

\usepackage{algorithm}
\usepackage[noend]{algorithmic}

\usepackage{tikz}
\usepackage{tkz-euclide}
\usepackage{csquotes}
\usepackage{booktabs}
\usepackage{caption}

\newcommand{\ignore}[1]{}

\newcommand{\xbest}{x^{\text{best}}}

\begin{document}

\title{One-Shot Decision-Making with and without Surrogates}

\author{Jakob Bossek\institute{The University of Adelaide,
Adelaide, Australia} %, \newline{}email: jakob.bossek@adelaide.edu.au}
\and Pascal Kerschke\institute{University of M{\"u}nster,
M{\"u}nster, Germany} %, \newline{}email: pascal.kerschke@wi.uni-muenster.de}
\and Aneta Neumann$^1$
%\institute{University of Adelaide, Adelaide, Australia}%, \newline{}email: aneta.neumann@adelaide.edu.au}
\and Frank Neumann$^1$ 
%\institute{University of Adelaide,Adelaide, Australia, \newline{}email: frank.neumann@adelaide.edu.au}
\and Carola Doerr\institute{Sorbonne University, CNRS, Paris, France}%,Paris, France, email: \newline{}carola.doerr@mpi-inf.mpg.de}
}

\maketitle
\bibliographystyle{ecai}

\begin{abstract}
One-shot decision making is required in situations in which we can evaluate a fixed number of solution candidates but do not have any possibility for further, adaptive sampling. Such settings are frequently encountered in neural network design, hyper-parameter optimization, and many simulation-based real-world optimization tasks, in which evaluations are costly and time sparse. 

It seems intuitive that well-distributed samples should be more meaningful in one-shot decision making settings than uniform or grid-based samples, since they show a better coverage of the decision space. In practice, quasi-random designs such as Latin Hypercube Samples and low-discrepancy point sets form indeed the state of the art, as confirmed by a number of recent studies and competitions. 

In this work we take a closer look into the correlation between the distribution of the quasi-random designs and their performance in one-shot decision making tasks, with the goal to investigate whether the assumed correlation between uniform distribution and performance can be confirmed. We study three different decision tasks: classic one-shot optimization (only the best sample matters), one-shot optimization with surrogates (allowing to use surrogate models for selecting a design that need not necessarily be one of the evaluated samples), and one-shot regression (i.e., function approximation, with minimization of mean squared error as objective). Our results confirm an advantage of low-discrepancy designs for all three settings. The overall correlation, however, is rather weak. 

We complement our study by evolving problem-specific samples that show significantly better performance for the regression task than the standard approaches based on low-discrepancy sequences. A cross-validation of these designs shows surprisingly good performance across the whole benchmark set, giving strong indication that significant performance gains over state-of-the-art one-shot sampling techniques are possible.  

Our results raise the important question which other diversity measures should be taken into account when designing distributions for one-shot decision making tasks. 
%This a problem that is highly relevant also in experimental design -- a problem closely related to one-shot decision making.     
%
%In this work we extend one-shot decision making by allowing exactly one adaptive evaluation. This process captures situations in which a final decision has to be made after an initial evaluation of some preliminary designs which are used to build a surrogate that models the function to be minimized. We empirically analyze five different experimental designs (four quasi-random vs. uniform sampling) and four different surrogate models (Random Forests, Kriging, SVMs, Decision Trees). We test their potential on the 24 noiseless, four-dimensional functions from the BBOB benchmark set. Our results confirm that the dominance of quasi-random designs over uniform ones extends to our surrogate-based decision making setting.  The discrepancy-accuracy correlation, however, is weaker than expected. We therefore also compare the regression accuracy of our five experimental designs with that of point sets that are individually designed for each of the 24 BBOB functions. These specifically designed samples increase the prediction accuracy substantially, but show significantly worse performance in the cross-validation for the other 23 functions of the benchmark set. 
\end{abstract}

% \jakob{Check consistency: COCO vs. BBOB, one-shot vs. oneshot, Fig. vs. Figure, Tab. vs. Table, ...}\carola{ich bin fuer oneshot, Fig., Table, COCO (ich meine BBOB ist nur der Wokshop, oder?). Bei den Sections ist doof, dass die keine Nummer haben, ggf sollte man darauf versichten auf "Section 4" zu verweisen... }\pascal{bbob (black-box optimization benchmark) ist die collection der probleme und coco (COmparing Continuous Optimizers) ist die darauf aufbauende software -- somit bin ich eher fuer bbob (koennen coco ja trotzdem 1x erwaehnen/zitieren). bei oneshot vs. one-shot ist es mir relativ egal, allerdings ist der Titel des papers derzeit one-shot ;-) figure/table vs. fig./tab. wuerde ich gerne einheitlich halten (entweder beides abkuerzen oder beides ausschreiben).. und bei den sections stimme ich zu - ohne nummerierung ist ein querverweis irgendwie nicht schoen}
% \jakob{Ich hab den bestehenden Text mal angepasst wie folgt:}
% \begin{itemize}
%     \item Referenzen abkürzen: Tab., Fig.
%     \item one-shot (wegen Titel)
%     \item BBOB statt COCO
% \end{itemize}

\section{Introduction}
When dealing with costly to evaluate problems under high time pressure, a decision maker is often left with the only option of evaluating a few possible decisions in parallel, in the hope that one of them proves to be a reasonable alternative. The problem of designing strategies that guarantee a fair chance of finding a good solution is studied under the term \emph{one-shot decision-making}. This research area has recently gained momentum in the context of training (deep) neural networks~\cite{BousquetGKTV17,nevergrad}, where a common approach is to train several networks in parallel and keeping only the best ones for further use. Other popular applications of one-shot optimization are in hyper-parameter tuning~\cite{BergstraB12,Teytaud19}. Several one-shot decision-making tasks are studied in the literature, with the most important ones being
\begin{itemize}
    \item \emph{classic one-shot optimization:} we evaluate a set of $n$ points and the best one of these is our decision
    \item \emph{one-shot optimization with surrogates:} we use the $n$ evaluated points to derive a decision $\hat{x}$, which is not necessarily one of the evaluated samples. 
    \item \emph{one-shot regression:} we use the $n$ evaluated samples to build an approximation $\hat{f}$ of the actual, unknown, function $f$, with the objective to minimize the difference between $\hat{f}$ and $f$, measured here in terms of mean-squared error.  
\end{itemize}
Tab.~\ref{tab:sum} (see next page) summarizes these settings and the considered performance measures. 

Several works, in particular the one of Bousquet et al.~\cite{BousquetGKTV17} and the more recent work by Cauwet et al.~\cite{Teytaud19}, show that quasi-random designs of low discrepancy are more suitable for the one-shot optimization task than i.i.d.~uniform samples or grid search. The overall recommendation propagated in~\cite{BousquetGKTV17} are randomly scrambled Hammersley point sets with random shifts. Other low-discrepancy point sets also perform well in their experiments, leaving us with the question if there is a correlation between the discrepancy of a point set and its performance in one-shot optimization. If such a correlation existed, one could hope to find even better one-shot designs by searching for point sets of small discrepancy --- a problem that is much easier (yet very hard) to address than the original one-shot optimization problem. We attack this question by comparing five different experimental designs -- three generalized Halton point sets, Latin Hypercube Samples (LHS)~\cite{LHS}, and randomly drawn points (see the section on low discrepancy sets below for more details) -- on the 24 noiseless BBOB functions~\cite{Hansen2009_Noiseless}, a standard test bed for black-box optimization.

\subsection{Summary of Our Results}
% Our first set of experiments studies the classic one-shot scenario described above, where the goal is to minimize the regret $\min_{i=1,\ldots, n}f(x^i)-f^*$ (note here that we use minimization as objective. All formal definitions will be given in the subsequent sections).
Our first set of experiments studies the \textbf{classic one-shot optimization} scenario described above, where the goal is to minimize the regret $|f(\xbest)-f^*|$ of the best found sample. While our experiments confirm the superiority of low-discrepancy point sets over random sampling, no clear correlation could be identified between the uniformity of their distribution -- measured in terms of the \emph{star discrepancy} -- and their performance. This refutes our hope that point sets with optimized discrepancy values could substantially boost performance in one-shot optimization.

\begin{table}
\begin{center}
\captionsetup{justification=centerlast}
{\caption{Summary of the three different one-shot settings investigated in this work, and their main performance measures. Abbreviations: $f^*=\inf_x f(x)$, MSE=mean squared error (here evaluated for 10\,000 i.i.d.~uniform samples).}
\label{tab:sum}}
\begin{tabular}{c|cc}
\toprule
                        & without surrogate & with surrogate\\
\midrule
         optimization &  $|f(\xbest)-f^*|$ & $|f(\hat{x})-f^*|$\\
         regression & --- & MSE\\
\bottomrule
\end{tabular}
\end{center}
\end{table}

We then extend the classic one-shot optimization setting in two ways: by adding one adaptive evaluation and by shifting the focus to regression rather than optimization. For both these scenarios the performance does not depend anymore on one particular point only, but on the combined information obtained through the evaluated samples. In these settings it could therefore be even more crucial to ensure that the samples are evenly distributed. %Tab.~\ref{tab:sum} summarizes the three settings considered in this work, which we describe in more detail in the remainder of this introduction. 

Our first modification is motivated by situations in which the one-shot evaluations mainly have an informative character, based on which a final solution $\hat{x}$ will be proposed. Put differently, we use the $n$ evaluated samples to build a surrogate model $\hat{f}$ modeling the true problem $f$. Using $\hat{f}$ we derive a point $\hat{x}$ which we hope to have decent quality. In the \textbf{one-shot optimization with surrogate}-setting the point $\hat{x}$ is assumed to be implemented, so that the quality of the so-derived decision is measured by the regret $|f(\hat{x})-f^*|$. 
%A surprising theoretical result on one-shot optimization with surrogates can be found in~\cite{Teytaud19}. The authors prove that, unless $n$ is very large, setting $\hat{x}=(0,...,0)$ is superior to setting $\hat{x}:=\xbest$ even if both the optimum and the samples are sampled from the same Gaussian distribution

One-shot optimization with surrogates is in particular relevant for applications in which the initial samples $x^1,\ldots, x^n$ are only partially evaluated (e.g., using some simulation or evaluating them on few instances only), whereas the full evaluation of $\hat{x}$ is very costly (in real-world applications it may even be infeasible to evaluate a point entirely). Our results demonstrate that problems of smooth structure tend to benefit quite significantly from the surrogate-assisted optimization. For other functions, in particular problems lacking global structure or being highly rugged/multimodal the surrogates can be deceptive and suggest points that can be much worse than the best design point. No clear winning pair of design and surrogate model can be identified, suggesting that a priori knowledge about the problem at hand can be quite crucial for selecting the best one-shot optimization technique. 

We finally change our focus from optimization to \textbf{one-shot regression}. That is, instead of aiming to find a single good decision, we now consider the ability of the different designs to allow for surrogates $\hat{f}$ that model the true problem as accurately as possible. More precisely, we aim to construct a model $\hat{f}$ that has a small mean-squared error (MSE) for uniformly sampled test points. 

In the one-shot regression model we not only compare the five designs considered in the optimization settings, but also evolve problem-specific designs that exhibit small MSE values. While these designs were originally only constructed to obtain a baseline against which we can compare our \mbox{(quasi-)random} designs, a cross-validation of the evolved designs on the other benchmark problems reveals surprisingly fine overall performance. The strategy to evolve these points might therefore be appealing in its own right. However, yet again, we observe that there is no clear winning design, nor any obvious correlation between discrepancy and performance in the one-shot regression problem. 

While our results answer the original question about the correlation between discrepancy and performance in one-shot decision making problems negatively, they reveal a clear need and may pave a way for identifying other diversity measures correlating well with performance in the one-shot settings. The evolved designs clearly indicate that we can expect to see significant performance gains from such an investigation. 
%Our long-term goal are the design of problem-aware selection mechanisms for the one-shot decision tasks summarized in Tab.~\ref{tab:sum}, and the evolved designs . %The recent work in~\cite{Teytaud19} (which only studies one-shot optimization). 

\subsection{Connections to Design of Experiment and Model-Based Optimization}
\label{subsec:connections}
It is likely that advances in one-shot decision making will also contribute to two related tasks: model-based optimization and design of experiments (DoE). 

Sequential model-based optimization (SMBO)~\cite{HWBWB2015}, also studied under the notion of global optimization or surrogate-based optimization, is a sequential method for global optimization of black-box functions where function evaluations are computationally expensive, e.g., require complex numerical simulations or actual physical experiments. In SMBO one uses the evaluated samples of an initial design to build a model of the true objective function, which is computationally fast or at least much faster to evaluate than the true objective function. In a sequential process this initial design is augmented by injecting further design points in order to improve the function approximation. So-called infill criteria or acquisition functions which usually balance exploitation of the current model and exploration of areas with high model uncertainty are used to decide which point(s) seem(s) adequate to evaluate next with the true objective function, i.e., how to trade a reduction of the uncertainty against a possible improvement. Classic model-based approaches, such as the efficient global optimization algorithm (EGO) by Jones et al.~\cite{EGO}, typically use well-distributed, space-filling point sets to initialize the search (see, e.g., \cite{FSK2008}). 

Also related to one-shot optimization is the classical Design of Experiments (DoE) scenario (see, e.g., the book by Box et al.~\cite{BHH2005}). Here, the goal is to find a good experimental design to explain variation in the data at hand or to get a good model fit to the data given a space-filling sampling plan, e.g., a Latin Hypercube Sample (LHS)~\cite{LHS}. These space-filling and ideally informative designs (see, e.g., \cite{wessing2015}) have shown to be an important ingredient when performing (automated) feature-based algorithm selection~\cite{kerschke2019bbob,kerschke2019flacco}.

Note though that despite the apparent similarity of these problems, the overall objectives are quite different, and good one-shot optimizers are therefore not necessarily good experimental designs nor good initialization rules for model-based optimization, and vice versa. Exploring connections between these three settings forms an interesting line for future research, which, however, we have to leave for future work. 
%As far as we know no tight connection between the scenarios 
%both these scenarios differ fundamentally from the one-shot settings discussed in this present work. While in the classical DoE setting, design points should be spread well across the search domain, in the one-shot setting a good coverage of the search space close to local or at best global optima is desired and sufficient, i.e., a good global search space coverage is not required. Good one-shot optimizers are therefore not necessarily good experimental designs, and vice versa.

\subsection{Reproducibility}
Upon acceptance of this ECAI submission, all experimental data of our study will be made available in the public GitHub repository \url{https://github.com/jakobbossek/oneshot/}.

%%%%%%%%%%%%%%%%%%
\section{Low-Discrepancy Designs}
\label{sec:discrepancy}

The discrepancy of a point set measures how far it deviates from a perfectly distributed set. Various discrepancy measures exist, providing different performance guarantees in quasi-Monte Carlo integration and other applications~\cite{BC87,KuipersN74,Mat99}. The arguably most commonly discrepancy metric is the \emph{star discrepancy}, which measures the largest absolute difference between the volume $V_y$ of any origin-anchored box $\smash{[0,y]:=\prod_{i=1}^d[0,y_i]}$ and the fraction of points contained in this box. That is, the star discrepancy of a point set $\{x^1,\ldots,x^n\} \in [0,1]^d$ is defined as 
\begin{align*}
    D^*(X) := \sup_{y \in [0,1]^d} \left| V_y - \frac{|[0,y]\cap X|}{n} \right|.
\end{align*}
%A low star discrepancy thus implies that the volume of each anchored box $[0,y]$ can be approximated by the fraction of points that it contains. 
%In other words, a lower discrepancy corresponds to a more even distribution of the points.

\begin{table}
\begin{center}
\captionsetup{justification=centerlast}
{\caption{Discrepancy value of the best design and the relative overhead of the other designs. Values for LHS, UNIFORM (UNIF.), and EVOLVED (EVOL.) designs are averaged.}
\label{tab:discrepancies}}
\footnotesize 
\begin{tabular}{r|r|rrrrr}
\toprule
\normalsize $n$ 	&  \normalsize Best & \normalsize BW	&	\normalsize Halton	&	\normalsize LHS	&	\normalsize UNIF.	& \normalsize EVOL.\\
\midrule
125	    &	0.056 &		12\%	&	48\%	&	49\%	&	185\%	& 156\%\\
1000	&	0.013 &		20\%	&	35\%	&	109\%	&	316\% & 343\%	\\
2500	&	0.008 &		0\%	&	6\%	&	295\%	& 371\%	& --	\\
5000	&	0.005 &		2\%	&	6\%	&	376\%	& 413\%	&	-- \\
\bottomrule
\end{tabular}
\normalsize
\end{center}
\end{table}

Low-discrepancy designs provide a proven guarantee on their asymptotic discrepancy value. They are a well-studied object in numerical analysis, because of the good error guarantees that they provide for numerical integration. The interested reader is referred to the survey~\cite{doerr2014calculation}, which covers in particular the computational aspects of star discrepancies relevant to our work. %The interested reader can find further details in~\cite{DoerrGW17} with further background and a survey covering computational aspects of star discrepancies. 
In our experiments we consider four different designs of low discrepancy, and we compare them to uniform sampling. More precisely, we study Latin Hypercube Samples (LHS; cf. McKay et al.~\cite{LHS}) and three variants of so-called Halton sequences: the original one suggested by Halton~\cite{Halton64}, an improved version introduced by Braaten and Wellter~\cite{Braaten1979}, and a third design which -- for the two sample sizes $n=125$ and $n=1\,000$ -- we obtain from a full enumeration and evaluation of all generalized Halton sequences. For our four-dimensional setting, these are $34\,560$ different designs each. Those are evaluated using the algorithm by Dobkin et al.~\cite{DEM96}, which has running time $n^{1+d/2}$. This exact approach becomes infeasible for larger sample sizes, and we use the best generator for $n=1\,000$ instead. The so-minimized Halton designs are referred to as ``Best'' in the remainder of this work. 
The discrepancy values of the different designs used in this paper are summarized in Table~\ref{tab:discrepancies}.

\begin{figure*}[!t]
    \centering
    \includegraphics[width=\textwidth, trim = 1mm 1mm 1mm 1mm, clip]{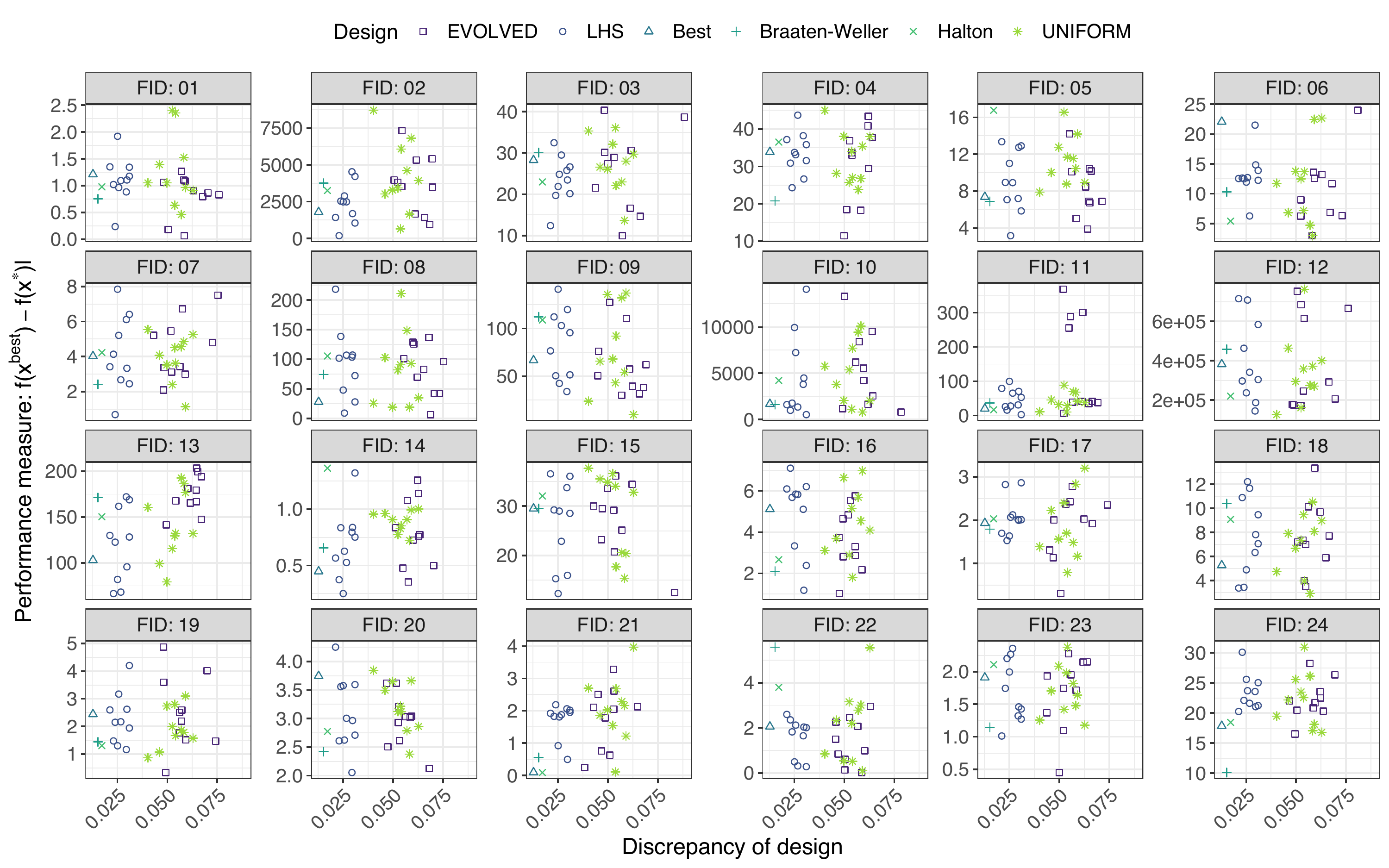}
    \captionsetup{justification=centerlast}
    \caption{Scatterplots showing the relationship between the discrepancy of designs of size $n = 1\,000$ and the one-shot performance $f(\xbest) - f^*$ %(difference of function values of best design point and known optimum) 
    for all 24 BBOB problems. The EVOLVED instances were designed for Kriging surrogates.}
    \label{fig:scatter_discrepancy_performance}
\end{figure*}

\section{Experimental Setup and Availability of Data} 

For our experiments we have chosen the 24 noiseless problems from the \emph{black-box optimization benchmark (BBOB)} by Hansen et al.~\cite{Hansen2009_Noiseless}, which is a well-established collection of continuous optimization problems. For computational reasons, we limit our attention to the first instance of each problem. The BBOB functions assume $[-5,5]^d$ as search space. % with $d$ being the dimension. 
We therefore scale our designs, which are initially constructed in $[0,1]^d$, accordingly. In the two optimization scenarios, we consider minimization as objective.  
 
%\begin{itemize}
%\item 24 bbob problems
%\item 5 sizes for the designs: 125, 1000, 2500, 5000, 10000
%\item 4 learner (except for n.points = 10000, where km was to expensive)
%\item three halton sequences (Best, BW, Halton) + 10x LHS + 10x Random + 10x Evolved (for the first 2 sample sizes)
% \end{itemize}

%\jakob{in total 124\,080, but we didn't introduce evolved at this point, so I will ignore it for now}\carola{I also mention them above, so that I would include them here as well. We will anyway have to mention them in the intro, so this should be OK.}
Our study summarizes the results from a total of $124\,080$ scenarios. We considered designs of three (deterministic) Halton sequences, as well as LHS and random uniform samples. As the latter two are stochastic, we generated ten samples each to account for their stochasticity. Furthermore, each design was generated for five different sample sizes $n \in \{125, 1\,000, 2\,500, 5\,000, 10\,000\}$. For each of those designs, we then computed surrogates using the following four machine learning algorithms: support vector machines (SVM~\cite{Cortes1995}), Decision Trees~\cite{Breiman1984}, Random Forests~\cite{Breiman2001}, and Kriging~\cite{Chiles2018}. Note that the latter could not be computed on designs of size $10\,000$ due to memory issues. Also, as the considered algorithms are stochastic -- or at least contain stochastic elements within their R implementations -- we replicated all experiments ten times. In addition to these $104\,880$ scenarios we further evaluated a total number of $19\,200$ ``evolved'' designs, which will be described in the section about one-shot regression.   

%%%%%%%%%%%%%%%%%%
\section{Classic One-Shot Optimization}
\label{sec:optimization}

%\pascal{in dieser section sollten wir noch mal betonen, dass die ML algorithmen hier erstmal egal sind.. das koennte confusing sein, da sie zuvor im setup genannt wurden.}\carola{done}

%We compare the suitability of three different Halton designs, as well as LHS and uniform sampling for the classic one-shot optimization task, i.e., without making use of any surrogate-assisted decisions. The aim of this task is 
In the classic one-shot optimization scenario we are asked to provide a point set $\{x^1,\ldots,x^n\}$ for which the quality $f(\xbest)$ of the best point $\xbest :=\text{arg}\,\min_{x^i \in \{x^1,\ldots,x^n\}} f(x^i)$ is as good as possible. 

In line with the machine learning literature, where the one-shot problem originates from, we consider simple regret $f(\xbest)-f^*$ as performance measure, where $f^*$ denotes the best function value $\inf_x f(x)$. Of course, this measure requires that $f^{*}$ is known -- which usually is not the case for real-world applications, but luckily those values are known for BBOB~\cite{Hansen2009_Noiseless}. 
Minimizing simple regret is also the standard objective in other related domains, including evolutionary computation~\cite{hansen2016cocoplat}. Since this performance depends on only one point, the variance of the results can be tremendous, and it is therefore interesting to compare different designs over different sets of problems (and to perform several independent runs in case of the stochastic designs LHS and uniform sampling).

Fig.~\ref{fig:scatter_discrepancy_performance} compares the average regret for each pair of function and design, and plots the respective performance ($y$-axis) in dependence of the design's discrepancy ($x$-axis). Due to different scales of the problems, performances should not be compared across functions. As already mentioned before, the results for LHS, UNIFORM and EVOLVED sampling are based on ten independent designs. Note that the concept of the EVOLVED designs will be discussed in more detail later in this work, but we are already showing its results for completeness. 

\begin{figure}[t]
    \centering
        \includegraphics[width=\linewidth]{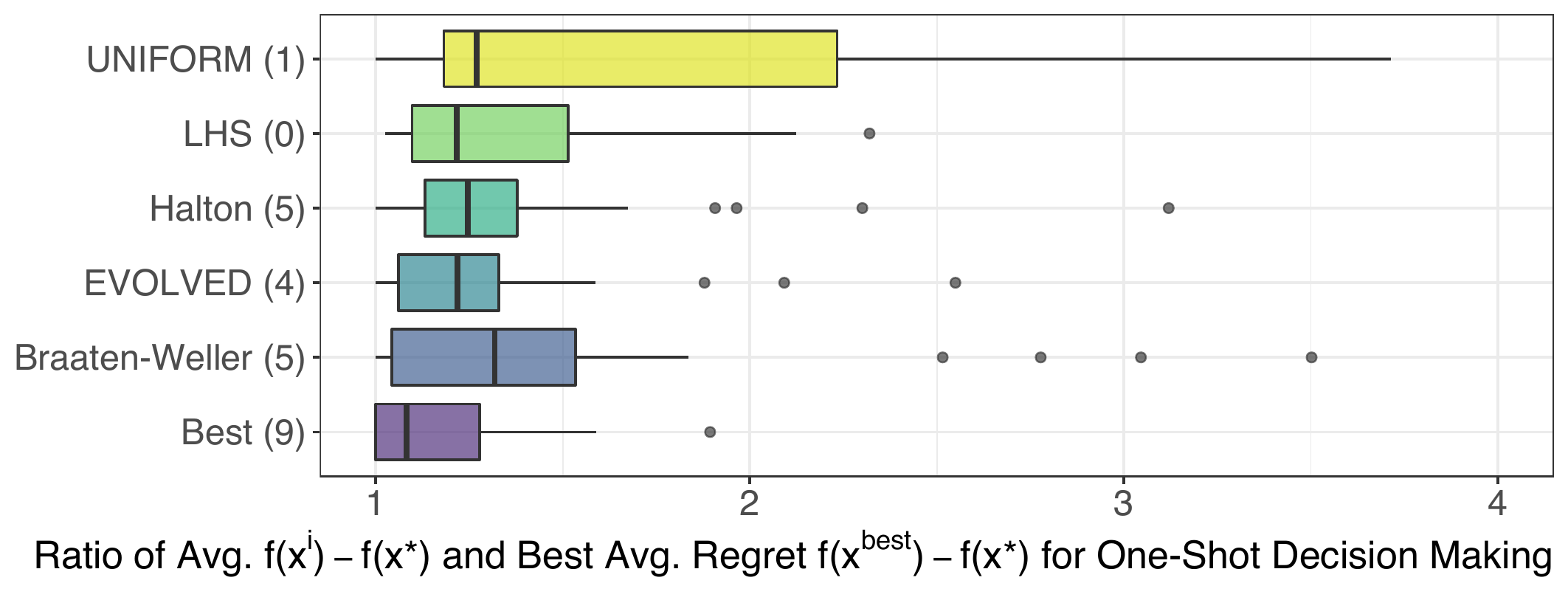}
        \captionsetup{justification=centerlast}
    \caption{
    Boxplots for factors by which the average one-shot result is worse than that of the best design (one data point per BBOB function).
    The $x$-axis is capped at 4 (outliers not shown in this plot: UNIFORM at 4.5 and 4.7, and Best at 7.2). Numbers in brackets indicate on how many functions the design achieved the best (average) result. All numbers are for $n=1\,000$ points and use Kriging surrogates.}
    \label{fig:oneshot-opt-boxplot}
\end{figure}

The plots in Fig.~\ref{fig:scatter_discrepancy_performance} indicate that the correlation between discrepancy and one-shot-performance is rather weak. However, we have to keep in mind that these performances depend on a single point only -- similar to a \emph{lucky punch} in sports. Therefore, we additionally analyze the aggregated performances in Fig.~\ref{fig:oneshot-opt-boxplot}. The boxplots display the distribution of the factor, by which each design is worse than the best design for the respective function. According to this aggregated view, the ``Best'' design -- whose discrepancy is the smallest among all sets (recall Tab.~\ref{tab:discrepancies}) -- is also the one achieving the smallest mean and median result. Further -- as indicated by the left border of the boxes -- the Braaten-Weller-design was highly competitive in 25\% of all cases. And although LHS showed good (average) performance as well, it failed to achieve the best average result on any of the functions. Interestingly, uniform sampling achieves a very good median score. In fact, we can see in Fig.~\ref{fig:scatter_discrepancy_performance} that the best uniform design often outperforms all other designs, but at the same time there is (with few exceptions) always at least one of the uniform samples which is worse than all other designs. 

%\carola{DEN PLOT HIER IN DIE SUPPLEMENTARY MATERIALS, WENN WIR WOLLEN:}
%\begin{figure*}[!t]
%    \centering
%    \includegraphics[width=\textwidth, trim = 1mm 1mm 1mm 1mm, clip]{plots-paper/lineplot_ranking.pdf}
%    \caption{Visualization of average ranks ($y$-axis; lower is better) w.r.t.~$|f(\xbest) - f^*|$ for each BBOB function ($x$-axis) split by learner and the method used to aggregate the performance of the uniform, LHS, and EVOLVED designs. The aggregation is either taking the \underline{min}imum, the \underline{mean} or the \underline{max}imun performance values with the first and last being optimistic and pessimistic perspectives respectively.}
%    \label{fig:lineplot_ranking}
%\end{figure*}

All in all, we learn from these results that discrepancy-minimization alone does not suffice to get good designs for one-shot optimization. However, we should keep in mind that, by choice, our benchmark problems are very diverse. In practice, we typically have some a priori information about the type of problem for which the one-shot optimization is asked for. It would therefore be very interesting to study if the rankings of the different designs are consistent for similar types of problems. To investigate this question, we suggest to not only consider more instances of the BBOB functions, but to extend our approach to other problems, such as those provided by Nevergrad~\cite{rapin2019,nevergrad} or the problems from the black-box optimization competition BBComp (\url{https://bbcomp.ini.rub.de}).  

%%%%%%%%%%%%%%%%%%
\section{One-Shot Optimization with Surrogates}
\label{sec:optwithsurrogates}

% \jakob{Wir sollten hier bzw. irgendwo im Paper mal SMBO (sequential model-based optimization)~\cite{EGO,Jones2001} erwaehnen und herausstellen, dass unser Ansatz im Wesentlichen ein Spezialfall ist: full exploitation des Surrogat-Models, da nur eine zusätzlich Evaluation erlaubt ist.}
We now consider an alternative one-shot optimization setting, in which the final decision does not have to be one of the $n$ evaluated points. That is, the decision maker is free to choose a final design $\hat{x}$ which is not identical to $\{x^1,\ldots,x^n\}$. This case covers in particular situations in which the evaluation of the $n$ points are only partial, e.g., because of a limited budget allocated to each evaluation or testing the setting only on some selected instances. Only one solution, $\hat{x}$, is then selected for a time-consuming, or otherwise costly, full evaluation. 
Note that this setting is loosely related to what is usually done in sequential model-based optimization (see Section~\ref{subsec:connections}). Our approach can be seen as a special case of SMBO with pure exploitation of the initial surrogate.

%\carola{Ich habe diesen Paragraph aus der Intro rausgestrichen, ggf sollte er mit dem Text unten gemerged werden. Auf jeden fall sollten wir die Referenzen uebernehmen.} The quality of the one-shot design depends on the point sets, the technique used to build the surrogate, and the method by which the local minimum $\hat{x}$ of the surrogate $\hat{f}$ is computed. In our empirical evaluation we consider the same five point sets as in the classical one-shot setting described above. We combine each of these five designs with four different surrogate models: support vector machines (SVM)~\cite{Cortes1995}, Decision Trees~\cite{Breiman1984}, Random Forests~\cite{Breiman2001}, and Kriging~\cite{Chiles2018}. 
%For the evaluation of the so-constructed surrogate, we use a standard local search approach, L-BFGS-B~\cite{Broyden1970}, which we starts in $\arg\min \{\hat{f}(x^i) \mid i\in [n]\}$. \carola{some text here about our findings}\pk{dafuer haben wir keinen platz mehr ;-)}\carola{das war ja nur der kopierte Absatz. Der ganze obere Absazt soll geloescht werden, wir brauchen nur ggf infos unten}

We would, a priori, expect that this freedom of selecting $\hat{x} \notin \{x^1,\ldots,x^n\}$ can only help the decision maker. However, note that choosing such a point also carries the risk that the quality $f(\hat{x})$ of the final solution is worse than $f(\xbest)$, the best of the already (partially) evaluated point. To investigate this trade-off between potential performance gains and the higher risk of choosing a point $\hat{x} \notin \{x^1,\ldots,x^n\}$, we analyze four different surrogate models: SVMs, Decision Trees, Random Forests, and Kriging. We feed each of these models with the design points $\{x^1,\ldots,x^n\}$ and their corresponding function values. Then a regression model $\hat{f}$ is build, which ideally provides a cheap copy of the underlying ``true'' function $f$. Assuming that $\hat{f}$ is indeed a good surrogate for $f$, one can now optimize $\hat{f}$ to find a good approximation for the optimum of $f$. For this purpose, we run the local search optimizer L-BFGS-B~\cite{Broyden1970} %BFGS~\cite{Broyden1970} 
on $\hat{f}$, starting in point $z = \text{arg}\,\min_{x^i \in \{x^1, \ldots, x^n\}} \hat{f}(x^i)$. The result of this local search will be a point $\hat{x}$, which is taken as the estimated optimum of function $f$. Our hope is to find that its true function value $f(\hat{x})$ is better than $f(\xbest)$. 

\begin{figure}[t]
    \centering
    \includegraphics[width=\linewidth, trim = 1mm 1mm 1mm 1mm, clip]{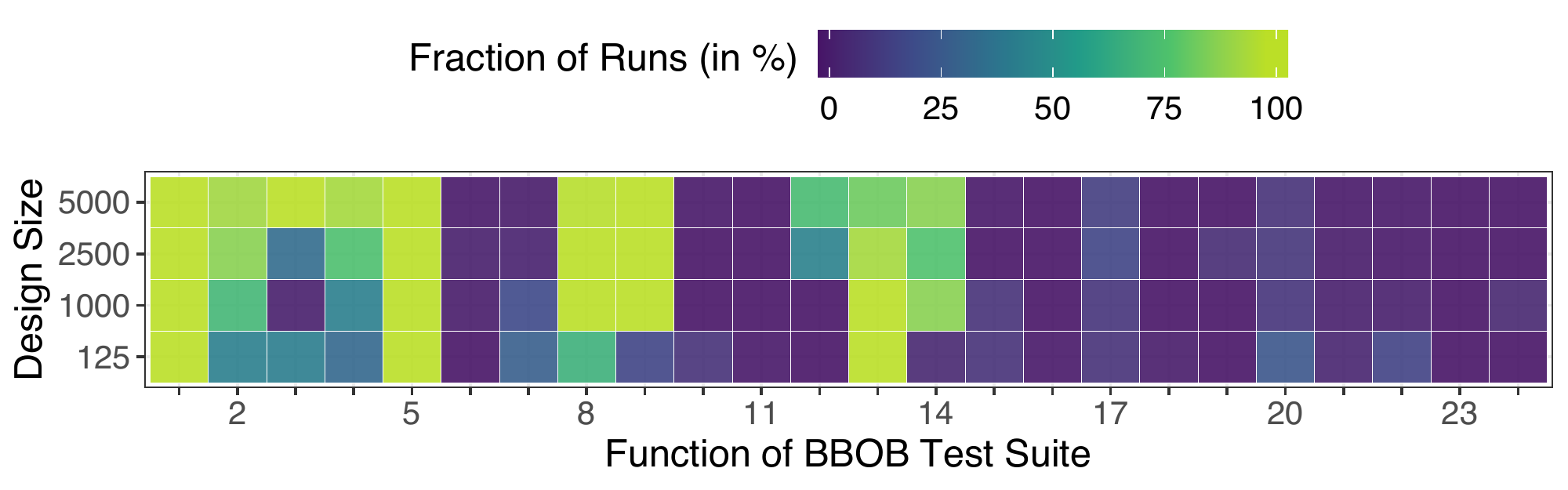}
    \captionsetup{justification=centerlast}
    \caption{Fraction of runs for which $f(\hat{x}) \le f(\xbest)$, by BBOB function (1-24) and sample size ($n=125$, $1\,000$, etc). All data is for LHS designs with Kriging surrogates.}
    \label{tab:oneshot-w-surrogates}
\end{figure}

Since some of the designs (LHS and UNIFORM sampling) and the surrogate models have random components, we take ten different designs and for each design ten independent runs of the surrogate model. Each of these runs will provide a different point $\hat{x}$. For each of these we compare $f(\hat{x})$ with $f(\xbest)$. Fig.~\ref{tab:oneshot-w-surrogates} shows the fraction of (design, repetition) pairs for which $\hat{x}$ is better than $\xbest$. Since we consider minimization, this is the case when $f(\hat{x})<f(\xbest)$. The results in Fig.~\ref{tab:oneshot-w-surrogates} are shown for the combination of LHS designs and Kriging models, and sorted by function and sample size. Each value is thus based on 100 different (design, repetition) pairs. 
For some functions we (almost) always find better points through the surrogate-based approach. In these cases, the advantages can be quite significant. For example,  
the average $f(\hat{x})$-value for function 5 and $n=1\,000$ points is a factor of 2.2 smaller than $f(\xbest)$. For other functions, however, we barely see any improvements and for some of the functions we even observe that $f(\hat{x})$ is much worse than $f(\xbest)$ -- the decision maker is thus deceived by the surrogate model. Looking into the structure of the 24 benchmark problems~\cite{Hansen2009_Noiseless}, we find that unimodal functions (FIDs 1 to 14) benefit quite significantly from the surrogate-based approach -- especially the separable functions 1 to 5. On the other hand, it can be observed that multimodal functions (15 to 24), and especially the ones lacking a global structure (20 to 24), are typically harmed by the surrogate-based approach. 

\begin{figure}[t]
    \centering
    \includegraphics[width=\linewidth]{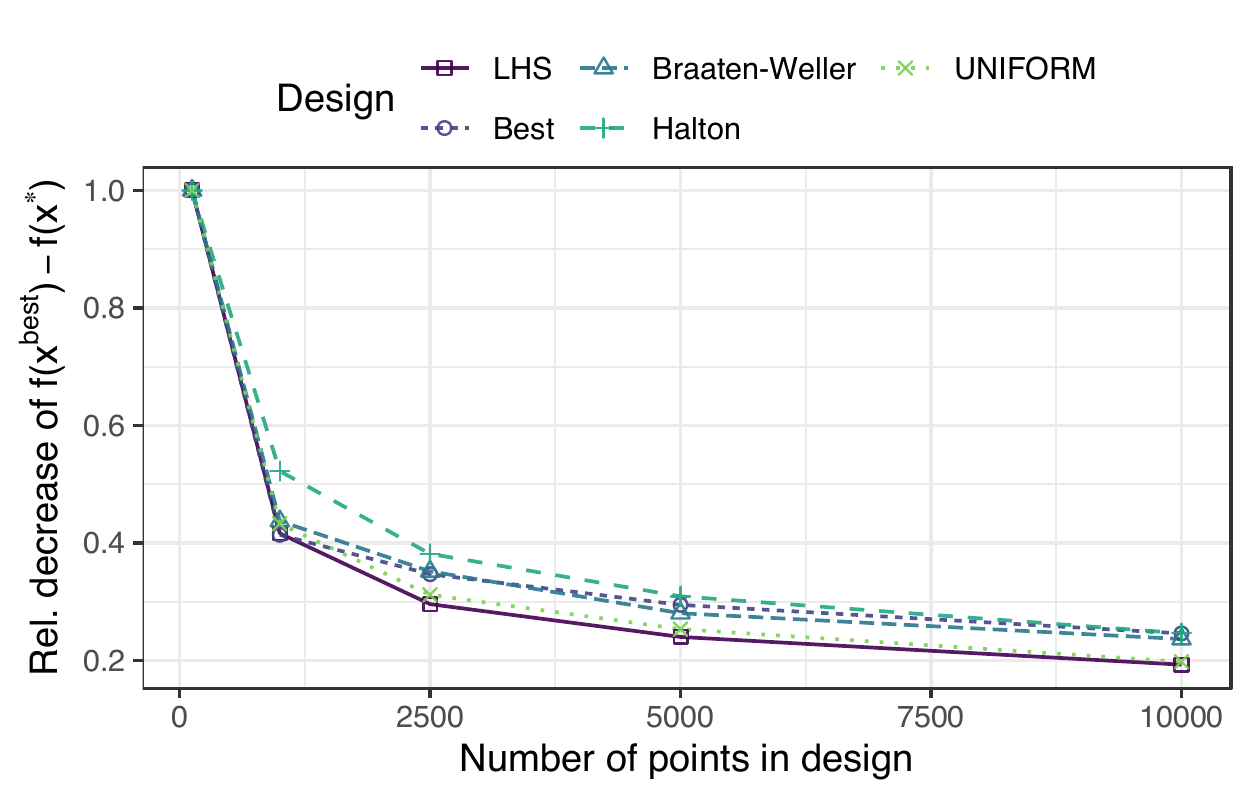}
    \captionsetup{justification=centerlast}
    \caption{Mean relative decrease (closer to zero is better) in one-shot performance $|f(\xbest) - f^*|$ with increasing sample size. Performance was scaled to $[0,1]$ for each function and design by normalizing with the worst one-shot performance obtained with $n=125$ points to allow for a fair comparison across all 24 BBOB functions.}
    \label{fig:n_vs_oneshot_performance}
\end{figure}

We further observe from Fig.~\ref{tab:oneshot-w-surrogates} that the advantage of the surrogate-based approach over classical one-shot optimization does not necessarily increase with increasing sample size. This may seem counter-intuitive at first, but we have to keep in mind that the quality of $f(\xbest)$ also increases quite drastically with the increasing number of samples $n$, see Fig.~\ref{fig:n_vs_oneshot_performance}. At the same time, the quality of the overall accuracy of the surrogate-based regression increases with increasing sample size (we will discuss this later during this work), so that the non-existing monotonicity of the data in Fig.~\ref{tab:oneshot-w-surrogates} should not be interpreted as a degeneration of the surrogates with larger samples.

% \ignore{
% \begin{table}[t]
% \centering
% \begin{tabular}{r|rrrr|r}
% FID & 125   & $1\,000$ & $2\,500$  & $5\,000$ & rel.gain ($n=5k$)  \\
%  \midrule
% 1  & 100\% & 100\% & 100\% & 100\% & 0\%   \\
% 2  & 45\%  & 72\%  & 89\%  & 94\%  & 105\% \\
% 3  & 44\%  & 3\%   & 36\%  & 100\% & 2\%   \\
% 4  & 34\%  & 45\%  & 77\%  & 95\%  & 3\%   \\
% 5  & 100\% & 100\% & 100\% & 100\% & 208\% \\
% 6  & 0\%   & 2\%   & 3\%   & 1\%   & 23\%  \\
% 7  & 30\%  & 20\%  & 4\%   & 2\%   & -1\%  \\
% 8  & 69\%  & 100\% & 100\% & 99\%  & 12\%  \\
% 9  & 18\%  & 100\% & 100\% & 100\% & 12\%  \\
% 10 & 10\%  & 0\%   & 0\%   & 1\%   & 22\%  \\
% 11 & 1\%   & 0\%   & 0\%   & 0\%   & 15\%  \\
% 12 & 0\%   & 0\%   & 46\%  & 75\%  & 58\%  \\
% 13 & 100\% & 100\% & 95\%  & 84\%  & 32\%  \\
% 14 & 6\%   & 89\%  & 78\%  & 89\%  & 0\%   \\
% 15 & 10\%  & 10\%  & 0\%   & 1\%   & -1\%  \\
% 16 & 1\%   & 0\%   & 0\%   & 0\%   & 0\%   \\
% 17 & 10\%  & 11\%  & 18\%  & 15\%  & 3\%   \\
% 18 & 2\%   & 0\%   & 0\%   & 0\%   & 3\%   \\
% 19 & 0\%   & 0\%   & 8\%   & 0\%   & 0\%   \\
% 20 & 26\%  & 11\%  & 12\%  & 10\%  & 0\%   \\
% 21 & 5\%   & 3\%   & 2\%   & 1\%   & 0\%   \\
% 22 & 17\%  & 3\%   & 1\%   & 1\%   & 0\%   \\
% 23 & 0\%   & 0\%   & 0\%   & 0\%   & 0\%   \\
% 24 & 0\%   & 6\%   & 0\%   & 0\%   & 0\%   \\
% \bottomrule
% \end{tabular}
% \caption{Fraction of runs for which $f(\hat{x}) \le f(\xbest)$, by BBOB function (1-24) and number of design points ($n=125$, $1\,000$, etc). All data is for Kriging surrogates and LHS designs.}
% \label{tab:oneshot-w-surrogates}
% \end{table}
% }

% \begin{table}[t]
% \centering
% \footnotesize
% \begin{tabular}{r|rrrr}
% \toprule
% \normalsize FID & \normalsize 125   & \normalsize $1\,000$ & \normalsize $2\,500$  \normalsize & $5\,000$  \\
%  \midrule
% 1  & 100\% & 100\% & 100\% & 100\%  \\
% 2  & 45\%  & 72\%  & 89\%  & 94\%    \\
% 3  & 44\%  & 3\%   & 36\%  & 100\%  \\
% 4  & 34\%  & 45\%  & 77\%  & 95\%    \\
% 5  & 100\% & 100\% & 100\% & 100\%   \\
% \midrule
% 6  & 0\%   & 2\%   & 3\%   & 1\%     \\
% 7  & 30\%  & 20\%  & 4\%   & 2\%    \\
% 8  & 69\%  & 100\% & 100\% & 99\%    \\
% 9  & 18\%  & 100\% & 100\% & 100\%   \\
% \midrule
% 10 & 10\%  & 0\%   & 0\%   & 1\%     \\
% 11 & 1\%   & 0\%   & 0\%   & 0\%     \\
% 12 & 0\%   & 0\%   & 46\%  & 75\%    \\
% 13 & 100\% & 100\% & 95\%  & 84\%   \\
% 14 & 6\%   & 89\%  & 78\%  & 89\%    \\
% \midrule
% 15 & 10\%  & 10\%  & 0\%   & 1\%     \\
% 16 & 1\%   & 0\%   & 0\%   & 0\%     \\
% 17 & 10\%  & 11\%  & 18\%  & 15\%    \\
% 18 & 2\%   & 0\%   & 0\%   & 0\%     \\
% 19 & 0\%   & 0\%   & 8\%   & 0\%     \\
% \midrule
% 20 & 26\%  & 11\%  & 12\%  & 10\%    \\
% 21 & 5\%   & 3\%   & 2\%   & 1\%     \\
% 22 & 17\%  & 3\%   & 1\%   & 1\%     \\
% 23 & 0\%   & 0\%   & 0\%   & 0\%     \\
% 24 & 0\%   & 6\%   & 0\%   & 0\%    \\
% \bottomrule
% \end{tabular}
% \normalsize 
% \caption{Fraction of runs for which $f(\hat{x}) \le f(\xbest)$, by BBOB function (1-24) and number of design points ($n=125$, $1\,000$, etc). All data is for Kriging surrogates and LHS designs.}
% \label{tab:oneshot-w-surrogates}
% \end{table}

Note that all our data is based on off-the-shelf surrogate models, i.e., all learners have been used in their default configurations. One could expect to see better results for tuned versions thereof. We consider an automated selection and configuration of the most suitable surrogate models a very promising research direction \cite{Saini2019}, which we intend to address in future works.% Fig.~\ref{fig:oneshot-by-surrogate} provides a first impression of the suitability of the different surrogate models. It compares the performance of $f(\hat{x})-f^*$ for the four different surrogate models, using the Braaten-Weller set with $n=1\,000$ points. \carola{NOCH UEBERLEGEN WAS MAN DA machen sollte: ranks? oder relative performance? JAKOB, PASCAL, Ideen?}\carola{fuer Journalversion dann...}

\begin{figure}[t]
    \centering
    \includegraphics[width=\linewidth, trim = 1mm 1mm 1mm 1mm, clip]{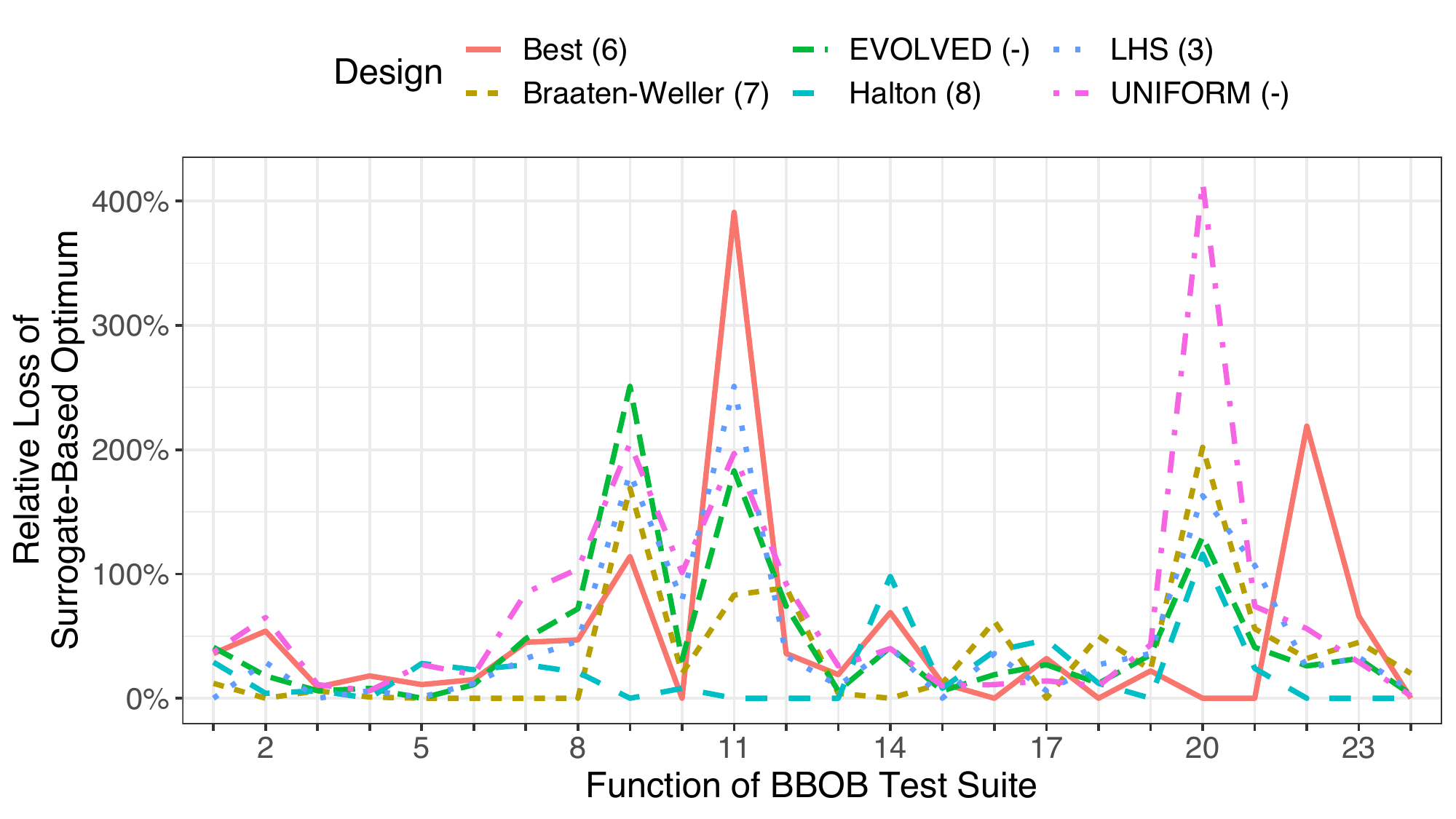}
    \captionsetup{justification=centerlast}
    \caption{Relative loss of estimated surrogate-based optimum against best estimate across all designs. Results are for $n=1\,000$ points and using Random Forest surrogates. Numbers in brackets indicate how often the design was best.}
    \label{fig:oneshot-w-surrogate}
\end{figure}

In Fig.~\ref{fig:oneshot-w-surrogate} we compare the quality of the different designs when using Random Forests as surrogate model and $n=1\,000$ points. We include in this plot the ``evolved'' point designs which will be described in Section~\ref{sec:evolve}. While those are never able to outperform all other designs in the surrogate-assisted  one-shot optimization setting, they still achieve fine performance for this task.  

% \begin{figure}[t]
%     \centering
%     \includegraphics[width=\linewidth]{plots-paper/opt-w-surrogates-RF-1000.png}
%     \caption{Relative loss of estimated surrogate-based optimum against best estimated across all designs. Results are for $n=1,000$ points and using Random Forest surrogates. Capped $y$-axis. Numbers in brackets indicate how often the design was the best one.}    \label{fig:oneshot-w-surrogate}
% \end{figure}

%%%%%%%%%%%%%%%%%%
\section{One-Shot Regression}
\label{sec:regression}

We now turn our attention to the \emph{one-shot regression} problem, in which we aim to build a regression model $\hat{f}$ that predicts the function values of the true function $f$ as accurately as possible. The accuracy of the \emph{one-shot regression models} is measured by the mean squared error (MSE), for which we evaluate both $f$ and our proxy $\hat{f}$ in $t$ i.i.d.~points $y^1,\ldots, y^t$, which are selected from the domain $[-5,5]^d$ uniformly at random. The MSE is then computed as 
$$
\text{MSE}(\hat{f}):= \frac{1}{t}\sum_{j=1}^t{\left(f(y^j)-\hat{f}(y^j)\right)^2}.
$$
Note that it seems likely that good experimental designs for the one-shot regression task, i.e., accurate global approximations of $f$ w.r.t.~MSE, are good initial samples for instantiating the sequential model-based optimization approaches mentioned in Section~\ref{subsec:connections}.

In Tab.~\ref{tab:winning-Kriging} we compare the different designs for different sample sizes. We count the number of functions for which the design achieved an MSE that is at most 5\% worse than the best one for the respective sample size. The displayed results are based on Kriging but results for the other surrogate models are similar.  %\carola{check..}
%We recall that there are a total number of 24 functions in our benchmark set. 
Uniform samples seem to enable less accurate regression models than the Halton and LHS designs. However, there are three cases in which the uniform design yields the best MSE: for function F16 with $n=125$ points, F22 with $n=1\,000$, and F3 with $n=2\,500$. In the latter case no other design achieves an MSE within the 5\% margin, whereas for the first two combinations the other designs achieve just slightly worse MSE-values.

\begin{figure*}[!t]
    \centering
    \includegraphics[width=\textwidth, trim = 1mm 2mm 1mm 1mm, clip]{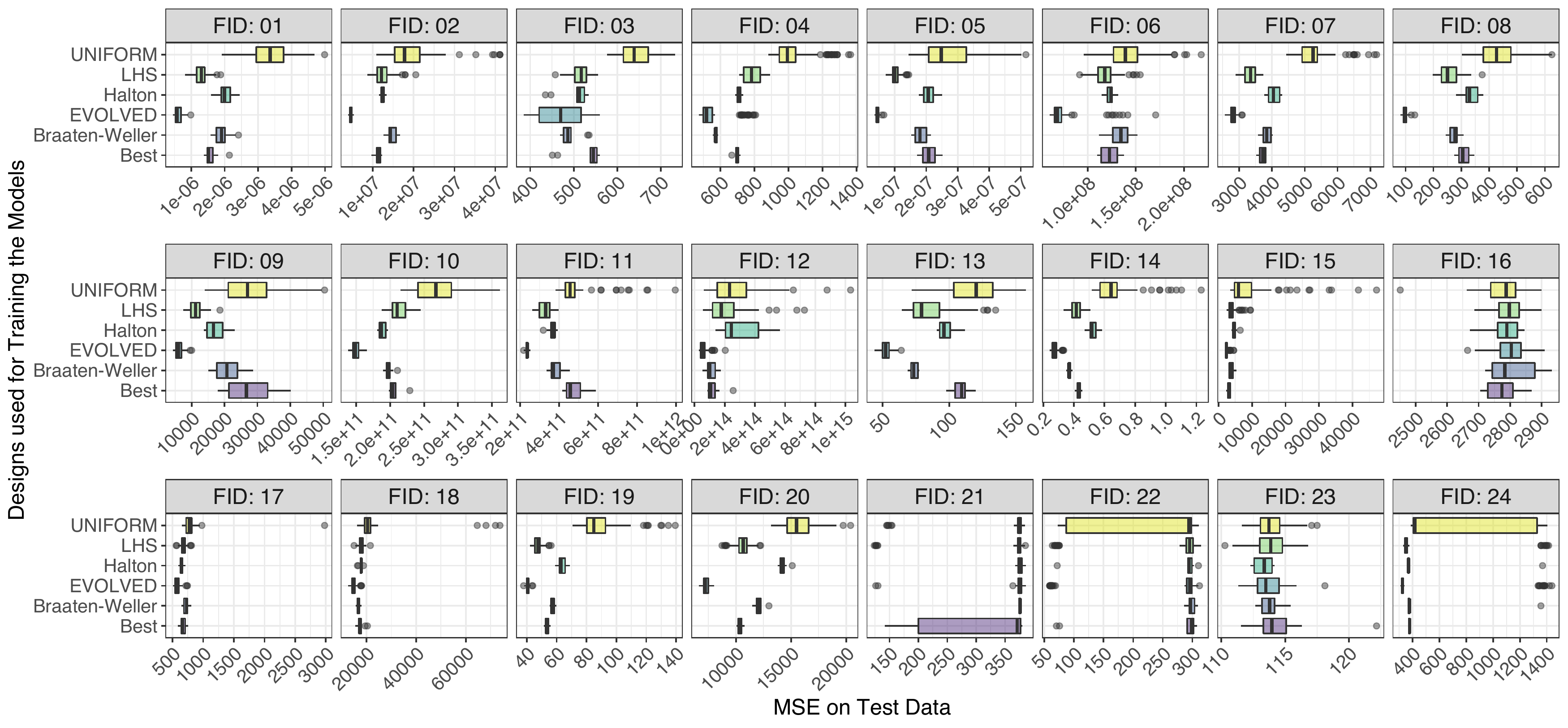}
    \captionsetup{justification=centerlast}
    \caption{Boxplots showing the MSE of the Kriging models across the six designs (used for training) and all 24 BBOB problems. Each model was trained using $n=1\,000$ points and afterwards assessed on a set of $10\,000$ random uniformly sampled points.}
    \label{fig:boxplot_mse_kriging}
\end{figure*}

\begin{table}
\begin{center}
\captionsetup{justification=centerlast}
{\caption{Number of functions for which the resp.~design, together with Kriging surrogates, achieved (on average) MSE that is at most 5\% worse than the best achieved MSE.}
\label{tab:winning-Kriging}}
\footnotesize 
\begin{tabular}{r|rrrrr|r}
\toprule
\normalsize $n$	&	\normalsize Best	&	\normalsize BW	&	\normalsize Halton	&	\normalsize LHS	&	\normalsize unif. & \normalsize total\\
\midrule
125	    &	6	&	13	&	8	&	11	&	4 & 42\\
$1\,000$	&	9	&	9	&	6	&	11	&	3 &  38\\
$2\,500$	&	11	&	7	&	8	&	16	&	3 & 45\\
$5\,000$	&	12	&	10	&	11	&	14	&	3 & 50\\
\bottomrule
\end{tabular}
\normalsize 
\end{center}
\end{table}

Fig.~\ref{fig:boxplot_mse_kriging} provides a more detailed impression of the regression quality for the different (design, function) pairs. This chart includes the EVOLVED designs, which we introduce and discuss in the next subsection. The results in Fig.~\ref{fig:boxplot_mse_kriging} are for Kriging, but those for the other models look alike. We observe clear patterns: uniform designs in general produce surrogate models with high mean MSE and high variance and hence a poor global approximation of the target function $f$ on average. In contrast, models fitted to LHS and low-discrepancy designs tend to be much more accurate approximations of the true function. However, there is no obvious winner, indicating that the correlation between discrepancy and performance is more complex than one might have hoped for.  

%%%%%%%%%%%%%%%%%%
\section{Evolving Designs for One-shot Regression}
\label{sec:evolve}

%\carola{hier faende ich es nett, wenn wir etwas mehr INfo zu den Ranks geben koennten. Wir hatten da mal einen Plot, aber den finde ich gerade nicht mehr (hattet Ihr in der AAAI submission Nacht gemacht)}

Given a target function $f$ and a surrogate model we do not know what quality -- w.r.t.~the MSE on test data -- one can achieve in the best-case with an optimal design of $n$ points in $d$ dimensions; a baseline is missing. In order to get an impression of the potential improvements we approximate optimal $n$-point designs by means of an evolutionary algorithm (EA; cf. Eiben and Smith~\cite{eiben2015}), i.e., we evolve sampling plans in a heuristically guided stochastic manner: First, we generate an initial LHS design $x$ of $n$ points in $d$ dimensions within the boundaries $[-5, 5]^d$. Subsequently, the EA performs small changes to a copy of $x$ by Gaussian perturbation of a subset of $\lfloor n/10 \rfloor$ points to create a mutant $y$.\footnote{Mutants violating the box-constraints are repaired by projecting the violating components to the boundary.} If $y$ is no worse w.r.t.~the fitness function, replace $x$ by $y$, otherwise discard $y$. The process is repeated for a fixed number of $2\,000$ iterations. The fitness function fits a surrogate model based on the given design in a first step. Next, the quality of the surrogate is assessed by means of the MSE for ten random uniform designs with $10\,000$ points each. The fitness value is the average of these MSE values and is meant to be minimized. Note that each run of the EA produces a large set of interim solutions, but we only keep the final design for further evaluation.

\begin{figure*}[!t]
    \centering
 %   \includegraphics[width=\textwidth, trim = 1mm 1mm 1mm 1mm, clip]{plots-paper/mse_matrix.pdf}
%    \caption{Illustration of MSEs across the 24 functions from the BBOB suite. The columns indicate for which problem the respective design has been optimized (w.r.t. the MSE) and the rows show to the corresponding test functions. The underlying surrogate models were trained using Kriging and designs of 125 points (left matrix) or 1000 points (right), respectively.}
   % \label{fig:mse_crossval}
    \includegraphics[width=\textwidth, trim = 1mm 1mm 1mm 1mm, clip]{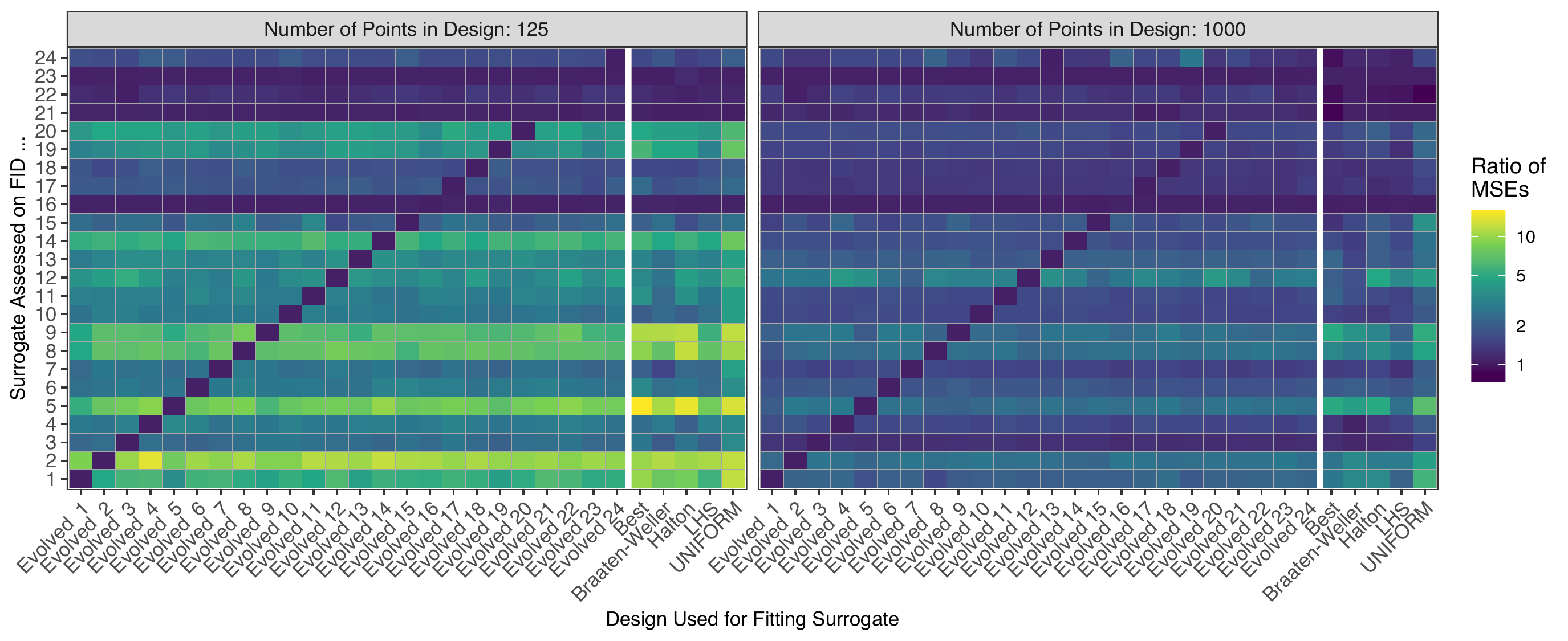}
    \captionsetup{justification=centerlast}
    \caption{Illustration of MSEs across the 24 functions from the BBOB suite. The first 24 columns correspond to the evolved designs (the $i$-th evolved design has been optimized for the $i$-th BBOB function), and the remaining five columns show the results for the five one-shot designs (Best, Braaten-Weller, Halton, LHS and UNIFORM). For each of the 29 designs, a Kriging model has been fitted to the BBOB function of the respective row and assessed by means of the MSE. The cell colors illustrate the ratio of the respective model's MSE and the MSE of the corresponding problem-tailored (i.e., evolved) design. %\carola{Erklaeren, was 1 ist --- MSE(von FID fuer FID, oder?)} \carola{hier waere es huebsch noch die SPalten fuer die anderen (oder \underline{ein} anderes) Design mit reinzuhmen. wenn nur 1s, dann zB LHS. Aber ist jetzt auch nicht sehr wichtig}\pascal{ist das jetzt verstaendlich beschrieben?}\jakob{Ich finde schon}\carola{yes, agree. Awesome!}
    }
    \label{fig:mse_matrix}
\end{figure*}

% \begin{figure}[t]
%     \centering
%     \begin{tikzpicture}[scale=0.45]
%         \input{tikz/ea_mutation.tex}
%     \end{tikzpicture}
%     \caption{Illustration of mutation  step of a design with $n = 10$ points (black dots) in two dimensions. Here, by example two points $x^i, x^j$ are subject to mutation (solid arrows). Note that perturbation of $x^j$ results in a point outside the bounding box. This is where a repair mechanism comes in (dashed arrow).}
%     \label{fig:ea_mutation}
% \end{figure}

% \begin{algorithm}[t]
%     \input{algorithms/evolving_ea.tex}
% \end{algorithm}

We evolved ten designs (to account for randomness of the EA approach) for each combination of surrogate modelling approach, BBOB function and sampling plan size $n \in \{125, 1\,000\}$ resulting in $1\,920$ EVOLVED designs. We neglected larger sampling sizes to keep computational costs reasonable.\footnote{Each fitness evaluation requires fitting a surrogate on $n$ points, which becomes computationally expensive for increasing $n$.} Moreover, we want to emphasize that the resulting designs are solely intended as \emph{baseline} -- a direct comparison with the other designs would thus be unfair as the problem-tailored designs evolved over numerous iterations and hence performed many more function evaluations.

Returning to Fig.~\ref{fig:boxplot_mse_kriging} we observe that the evolved designs lead to drastic improvements w.r.t.~the MSE (and low variance) for the majority of BBOB functions; in particular for FIDs 1-14 (first three BBOB groups with mainly unimodal functions with global structure). Contrary, for FIDs~16 and 21-24 -- functions which are characterized by a highly rugged landscape with many local optima and weak global structure -- the evolving process is far less successful w.r.t.~MSE improvement. These findings are in-line with our observations in the context of one-shot optimization with surrogates, i.e., if the function approximation is weak, so is likely the result of L-BFGS-B (or any alternative local-search algorithm) started from $\xbest$.

\begin{figure}[t]
\centering
\includegraphics[width=\columnwidth]{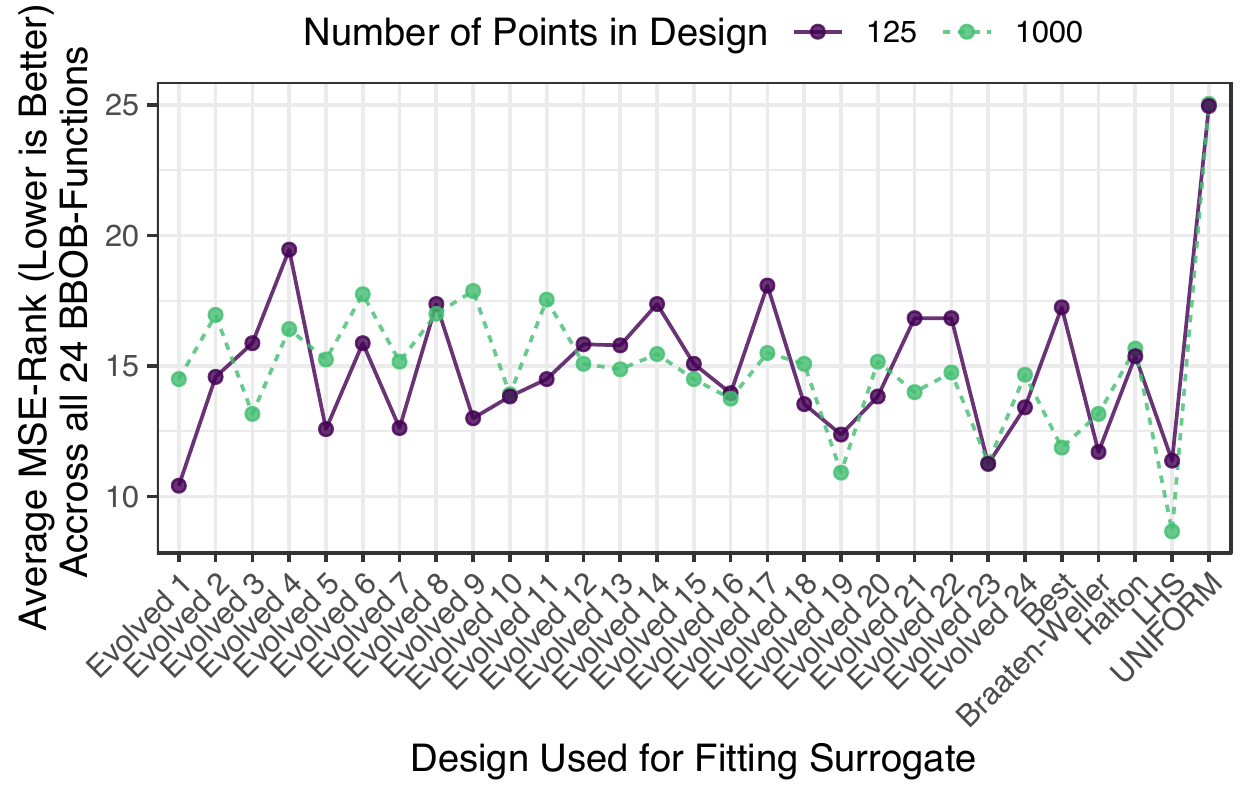}
\captionsetup{justification=centerlast}
\caption{Visualization of average MSE-ranks (lower is better) for all 29 designs. This figure aggregates the detailed MSE values in Fig.~\ref{fig:mse_matrix} across all 24 BBOB functions.}
\label{fig:mse_rank_lineplot}
\end{figure}
Recall that we evolved designs for specific combinations of target function, surrogate-modelling approach, and sample size. However, as depicted in Fig.~\ref{fig:mse_matrix}, the problem-specifically evolved designs are not necessarily inferior to any of the established sampling strategies. While the designs resulted indeed in significantly superior performances on the problems they have been evolved on -- as can be seen by the diagonal of dark blue cells -- their MSE ratios are usually comparable, if not even better, then the respective ratios of Best, Braaten-Weller, etc. When comparing the average ranks (see Fig.~\ref{fig:mse_rank_lineplot}) obtained by the 29 designs, the design evolved for BBOB function 1 achieves the best average score (10.4) of all 29 tested designs for $n=125$, closely followed by LHS (11.4) and Braaten-Weller (11.7). For $n=1\,000$ LHS scores on rank 8.7 on average, while the runner-ups are ``Evolved 23'' (11.3) and Best (11.9). Several other evolved designs obtain fine average ranks. The uniform design is clearly the worst, with average rank 25.0 for both $n=125$ and $n=1\,000$.  

Note though that we also observe from Fig.~\ref{fig:mse_matrix} quite noticeable differences across the functions on which the trained surrogates are assessed (rows). In addition, we observed a decrease in the MSE ratios for an increase in sample size. %This effect is absolutely plausible as larger samples enable more accurate surrogate models -- resulting in a shrinking MSE. \carola{hm, das erklaert ja nicht, warum die Ratios sinken, oder?}

\section{Conclusion}
\label{ces:conclusion}

We have analyzed the question whether the promising results of low-discrepancy point sets for one-shot optimization are well correlated with the discrepancy of these sets. Unfortunately, no strict one-to-one correlation could be identified, neither in the classic nor in the surrogate-assisted one-shot optimization setting, nor in the one-shot regression scenario. These results refute our hope that the challenging and resource-consuming task of designing efficient one-shot designs could be reduced to a discrepancy-minimization problem (which is also a challenging task in its own, see~\cite{DoerrR13,RainvilleGTL12}, but of a much smaller scale than the one-shot design one). It remains to investigate whether other diversity measures show a better correlation. Among the most promising candidates are indicators measuring how ``space-filling'' the designs are. Note though that good designs for one-shot \emph{optimization} need not necessarily be optimal also for one-shot \emph{regression}. This forms another research question to investigate. %Our data suggests that the general picture is similar: low-discrepancy designs outperform uniform samples, but when we reduce the focus to the high-performing designs, these analogies disappear. 
%Another question that we aim to attack in future work are the differences between good designs for one-shot \emph{optimization} vs. one-shot \emph{regression}. At the moment it is known in which way these two tasks require different designs. Our data suggests that the general picture is similar: low-discrepancy designs outperform uniform samples, but when we reduce the focus to the high-performing designs, these analogies disappear. 
%Although these two task do not necessarily require the same properties, we still observe consistently better performance of low-discrepancy designs vs. uniform samples. 
%However, despite a missing one-to-one correlation we nevertheless observe that certain space-filling properties do improve quite significantly over uniform designs, so that different diversity metrics might show a better correlation. 

The decent performance of the problem-specific designs obtained through our evolutionary approach was a big surprise. Not only did they improve quite considerably over the standard designs for one-shot regression for the problem and learner they were evolved for, but some of them even rank in the top places when evaluated across the whole benchmark set. A cross-validation of the evolutionary approach on other benchmarks and an extension to other dimensions forms another line of research that seems very promising in the context of one-shot decision making. 

%Finally, we have considered the one-shot decision making tasks as pure \emph{black-box optimization problems,} i.e., we did not take into account any problem-specific information when selecting the one-shot designs. In real-world scenarios we typically \emph{do} have some a priori knowledge about the problem at hand. Using such knowledge for an automated selection and configuration of the one-shot designs forms a third line of future work. Note that a problem-aware (aka \emph{gray-box optimization}) selection seems very useful also for the decision whether or not to use a surrogate-assisted optimization approach. As we have seen above (Fig.~\ref{tab:oneshot-w-surrogates}), some functions can benefit quite drastically from such an approach, while others may suffer considerably from deceptive surrogates. 
Finally, we consider the mentioned automation of configuring the one-shot designs through a problem-aware approach another promising perspective. Our data suggests that such a \emph{gray-box} design seems very useful for the decision whether or not to use a surrogate-assisted optimization approach. As we have seen in Fig.~\ref{tab:oneshot-w-surrogates}, some problems can benefit quite drastically from the adaptive choice, whereas others may suffer considerably from deceptive surrogates.

\ack This work has been supported by the Paris Ile-de-France region, and by a public grant as part of the
Investissement d'avenir project, reference ANR-11-LABX-0056-LMH,
LabEx LMH. Moreover, Pascal Kerschke acknowledges support from the \textit{European Research Center for Information Systems (ERCIS)}. This work has been supported by the Australian Research Council through grant DP190103894, DP160102401 and by the South Australian Government through the Research Consortium "Unlocking Complex Resources through Lean Processing",

%\bibliography{references}

\end{document}